# Generating Feasible and Diverse Synthetic Populations Using Diffusion Models

Min Tang, Peng Lu, and Qing Feng

*Abstract*—Population synthesis is a critical task that involves generating synthetic yet realistic representations of populations. It is a fundamental problem in agent-based modeling (ABM), which has become the standard to analyze intelligent transportation systems. The synthetic population serves as the primary input for ABM transportation simulation, with traveling agents represented by population members. However, when the number of attributes describing agents becomes large, survey data often cannot densely support the joint distribution of the attributes in the population due to the curse of dimensionality. This sparsity makes it difficult to accurately model and produce the population. Interestingly, deep generative models trained from available sample data can potentially synthesize possible attribute combinations that present in the actual population but do not exist in the sample data (called 'sampling zeros'). Nevertheless, this comes at the cost of falsely generating the infeasible attribute combinations that do not exist in the population (called 'structural zeros'). In this study, a novel diffusion model-based population synthesis method is proposed to estimate the underlying joint distribution of a population. This approach enables the recovery of numerous missing sampling zeros while keeping the generated structural zeros minimal. Our method is compared with other recently proposed approaches such as Variational Autoencoders (VAE) and Generative Adversarial Network (GAN) approaches, which have shown success in high dimensional tabular population synthesis. We assess the performance of the synthesized outputs using a range of metrics, including marginal distribution similarity, feasibility, and diversity. The results demonstrate that our proposed method outperforms previous approaches in achieving a better balance between the feasibility and diversity of the synthesized population.

*Index Terms*—Population synthesis, diffusion model, feasibility, diversity, agent-based modeling.

## I. INTRODUCTION

AGENT-BASED modeling (ABM) is a computational method that simulates complex systems by modeling individual agents and their interactions, enabling researchers to gain insights into system behavior. This approach has been widely applied in transportation systems [1], [2], [3], where the interactions between the mobility of traveling agents are studied. The key input of ABM is the synthetic population, which is used to characterizing the traveling agents. Since individual-level attributes of the entire population are unavailable, small-scale samples obtained through survey are generally used for population synthesis, i.e., creating realistic yet artificial population of agents based on a real sample of such agents [4]. In addition to privacy and confidentiality concerns, acquiring individual census survey data can also be time-consuming and resource-intensive, which limits its widespread adoption. Therefore, developing methods for generating synthetic populations from limited sample data is essential.

Population synthesis consists of three stages [5], [6], [7]: (1) generating a synthetic pools of individuals, (2) matrix fitting to ensure representative samples, and (3) allocating prototypical agents to micro-agents. This study focuses on the first stage of population synthesis, specifically generating synthetic pools of individuals that represent realistic and diverse combinations of individual attributes, which improves the overall accuracy of the ABM by circumventing the error propagation to later modelling stages.

Traditional sampling-based methods for population synthesis [7], [8], [9], have a significant limitation: they cannot generate sampling zeros, which is particularly problematic in scenarios involving small-scale training samples. In such cases, feature combinations sampled using re-weighting methods are even less representative of the population. To overcome this limitation, a promising solution is to employ a simulation-based approach leveraging generative models to produce synthetic pools [10], [11], [12], [13]. By doing so, the generative model first learns the joint probability distributions of the attributes, followed by sampling, which enables the generation of synthetic agents with attribute combinations that may not have been observed in the samples themselves, yet are likely present in the population.

Recently, the incorporation of deep generative models into population synthesis has led to the development of advanced methods for generating synthetic pools [6], [14], [15], [16]. Borysov et al. [6] introduced a variational autoencoder (VAE)-based deep generative model to produce synthetic micro-agents, while Garrido et al. [15] employed a Wasserstein Generative Adversarial Network (WGAN) for large-scale population synthesis. These models achieved promising results in reproducing the aggregated marginal and bivariate distribution of attributes. However, as Theis et al. [17] noted, such metrics can be inadequate and misleading in high-dimensional data, leading to overfitting the training data distribution [16]. Furthermore, these models have shown

This work was supported in part by the Central Guidance for Local Developmental Initiatives under Grant 2023EGA035 (China). *(Corresponding author: Min Tang.)*

M. Tang and Q. Feng are with the PKU-Wuhan Institute for Artificial Intelligence, Wuhan 430047, China (e-mail: gfkdtangmin@hotmail.com; fengqing@whai.pku.edu.cn).

P. Lu is with the PKU-Wuhan Institute for Artificial Intelligence, Wuhan 430074, China, and also with the Social Computing Center, Central South University, Changsha 410017, China (e-mail: lupengccps@vip.qq.com).

capability in generating sampling zeros at the cost of introducing structural zeros. Nevertheless, the feasibility and diversity metrics have not yet been sufficiently discussed in the context of population synthesis. Notably, a lower generation probability of structural zeros indicates higher feasibility of the generated data, whereas a lower generation probability of sampling zeros suggests lower diversity. Building on this understanding, Kim and Bansal [18] customized GANs and VAEs for synthetic pools generation and developed novel metrics for evaluating the feasibility and diversity of population synthesis.

Diffusion models [19], [20], [21], [22] have recently emerged as the state-of-the-art family of deep generative models, particularly exceling in approximating the full joint distribution of high-dimensional data sets. This breakthrough has far-reaching implications for a wide range of applications, including image generation, audio generation and video generation. However, the application of diffusion models to population synthesis has not been thoroughly explored.

This study aims to bridge this gap by investigating the feasibility and effectiveness of using diffusion models in population synthesis, with a focus on evaluating their performance in generating realistic and diverse populations.

In this paper, a novel method is presented for generating synthetic populations in agent based transportation simulation, achieved through the adaption of the denoising diffusion probabilistic model (DDPM) [21]. This approach enables the learning of the joint distribution of individual-level attributes from available sample data, facilitating realistic population synthesis. In summary, our contributions expand on the current literature on population synthesis as follows:

1) We propose a customized diffusion model to approximate the joint distribution of the high-dimensional population data, enabling more realistic and accurate synthetic populations generation in agent based transportation simulation. In the self-attention modules of our modified diffusion model, we replace the conventional 2D convolution operations in the self-attention modules with 1D convolutions, which are tailored to the continuous embeddings produced by the learnable mapping. This modification enables the effective exploration of intricate dependencies and relationships between diverse attribute variables.

2) By constructing a vocabulary consisting of all discrete attribute values, we devised a learnable mapping to transform the original discrete attributes into continuous embeddings, accommodating the continuous input condition of the DDPM.

3) We comprehensively compare and evaluate the performance of our proposed method for generating synthetic populations against state-of-the-art models, including VAEs and GANs, utilizing three critical evaluation metrics: distributional similarity, feasibility, and diversity.

The modified diffusion model can be trained on just 5% of the target population to produce a comparable number of synthetic data, demonstrating its efficiency and scalability. This breakthrough has significant implications for population synthesis applications, enabling the generation of high-quality synthetic pools with reduced computational burdens.

The paper is organized as follows. In section II, the literature review is presented. Section III formulates the problem and introduces our proposed methodology. In section IV, the evaluation results are discussed. Section V provides a conclusion and outlines future research directions.

## II. RELATED WORKS

### A. Population Synthesis

Population synthesis involves generating an artificial yet realistic population from an existing limited sample of real observations. The resulting synthetic population is a crucial input for agent-based transportation simulation [1], [2], [3]. A primary goal of any population synthesizer is to preserve the essential characteristics of the original population, driving the development of diverse technical approaches. These approaches can be broadly categorized into three types [23]: (1) re-weighting methods, which adjust the weight of existing data; (2) matrix fitting techniques, which optimize a set of parameters to match the target distribution; and (3) simulation-based approaches, which generate synthetic data through iterative modeling.

The re-weighting approach estimates expansion factors for a survey to adjust its sample to match the characteristics of the target population. This typically involves non-linear optimization methods [24], [25], but scalability issues may arise in high-dimensional scenarios. Matrix fitting extends re-weighting approaches by calculating expansion factors as ratios between the initial and target matrices. Methods like Iterative Proportion Fitting (IPF) [26] and maximum cross-entropy [27] are popular matrix fitting techniques. Related research has explored various aspects of matrix fitting, including convergence properties [28], the preservation of odds-ratios [29], the handling of convex constraints [30], [31], and the equivalence between cross-entropy and IPF [32]. Both matrix fitting and re-weighting produce samples of individuals with weighted characteristics, rather than true agent-based samples. Therefore, a post-simulation stage is necessary for population synthesis linked to agent-based models, where individuals are drawn from the weighted sample [7].

These methods belong to a family of strictly deterministic approaches that focus on adjusting sample data or matrices to match specified constraints or marginal distributions without incorporating probabilistic or stochastic elements. For low-dimensional problems, re-weighting and matrix fitting typically perform well and are not adversely affected by sparsity. However, these methods struggle to effectively approximate high-dimensional data structures due to their strict dependence on original sample data. Accurately modeling high-dimensional data often requires large samples that may be impractical or infeasible to obtain. A common problem when using these methods for higher dimensions is the inability to synthesize sampling zeros.

Simulation-based methods using generative models are gaining popularity due to their advantages over deterministic approaches. These models provide systematic ways to impute or interpolate data [10], [11], [12], [13]. One key benefit of generative models is their ability to learn the joint probability distribution of attributes, allowing them to generate specific, realistic combinations that may not have been present in the original data. This capability makes them particularly well-suited for tackling high-dimensional problems and creating detailed, accurate representations of populations.

Recently, deep generative models have been employed to generate synthetic populations [6], [14], [15], [16]. These models integrate deep neural networks into the generative process to learn complex patterns in real-world population data. Two widely-used deep generative models, the VAE [6] and the GAN [15] achieved promising performance in reproducing the aggregated marginal and bivariate distributions of attributes.

Kang et al. [33] proposed a diffusion model for population synthesis, which forcefully reshapes individual sample data into square matrices from vectors, formally aligning the individual data with the input format required by the original diffusion model designed for image generation. However, the representation capability of diffusion model for population data has not been fully explored. In evaluating the diffusion model, only the standardized root mean square errors (SRMSE) of marginal and bivariate distributions of attributes are calculated. These metrics often yield inadequate and misleading results when applied to high-dimensional data [17], potentially leading to overfitting of the simplified training data distribution as noted in [16]. Moreover, these metrics fail to assess the generative model's ability to accurately produce sampling zeros, a distinct strength of generative models.

In computer vision, feasibility (fidelity) and diversity metrics are used to evaluate image deep generative models. These metrics are closely tied to the probabilities of generating structural zeros and sampling zeros during synthetic population generation. Specifically, a lower probability of structural zeros indicates higher feasibility, while a lower probability of sampling zeros indicates lower diversity. Despite their importance in evaluating generative models, feasibility and diversity metrics have not been extensively explored in the context of population synthesis. A recent study by Kim and Bansal [18] addressed this gap. The authors introduced a comprehensive evaluation framework that utilizes large-scale data to assess feasibility and diversity in population synthesis using novel metrics related to structural and sampling zeros. We adopt the diversity and feasibility metrics for evaluating our proposed diffusion model based population synthesis method.

*B. Denoising Diffusion Probabilistic Models (DDPMs)*

Diffusion models are a class of probabilistic generative models that operate by progressively adding noise to data, then learn to reverse this process to generate samples. The success of Stable Diffusion [34] has drawn significant attention to diffusion models as promising deep generative models. In particular, the research on high-resolution image synthesis showcased in Stable Diffusion has highlighted the capabilities and advancements of generative modeling using diffusion models. There are three main categories of diffusion models: denoising diffusion probabilistic models (DDPMs), score-based generative models (SGMs), and stochastic differential equations (Score SDEs). We give a brief introduction to DDPM, which serves as the foundation for the diffusion model employed in this research.

A denoising diffusion probabilistic model (DDPM) [19], [21] employs two Markov chains: a forward chain that gradually adds noise to the data, and a reverse chain that learns to reconstruct the original data from this noise. The forward chain is typically designed to transform the complex data distribution into a simpler prior distribution, often a standard Gaussian. Conversely, the reverse Markov chain uses deep neural networks to learn transition kernels that reverse the effects of the forward chain. By doing so, it captures the intricate dependencies and patterns in the original data.

To generate new data points using DDPM, samples are first drawn from the prior distribution (e.g., a standard Gaussian). These samples are then processed through the reverse Markov chain via ancestral sampling [35], effectively reversing the noise-adding process to produce synthetic data that resembles the original distribution.

The processes of the DDPM model are depicted in Fig. 1 below.

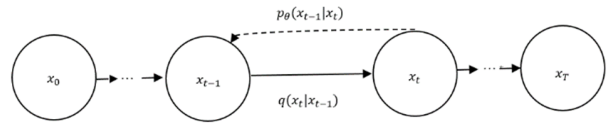

**Fig. 1.** The Markov chains of the forward and reverse diffusion processes.

Given a data distribution $x_0 \sim q(x_0)$, the forward Markov process generates a sequence of random vectors $x_1, x_2, \ldots, x_T$ with transition kernel $q(x_t|x_{t-1})$. The most common choice for this transition kernel is a Gaussian distribution, specifically:

$$q(x_t|x_{t-1}) = N(x_t; \sqrt{1-\beta_t}x_{t-1}, \beta_t \mathbf{I}) \quad (1)$$

The hyperparameters $\beta_t \in (0,1), t = 1, \ldots, T$ are chosen before training, where $\beta_t$ increases as $t$ becomes larger. As the step $t$ grows, the data sample $x_0$ gradually loses its distinct features. Eventually $x_T$ approximates an isotropic Gaussian distribution when $T$ is sufficiently large. Notably, the analytic form of $q(x_t|x_0)$ can be derived through iterative utilization of the equation (1) and the reparameterization trick.

$$q(x_t|x_0) = N(x_t; \sqrt{\bar{\alpha}_t}x_0, (1-\bar{\alpha}_t)\mathbf{I}) \quad (2)$$

where $\alpha_t = 1 - \beta_t$ and $\bar{\alpha}_t = \prod_{s=0}^{t}\alpha_s$. The reverse Markov chain, which is indeed a Markov chain as proven in [36], is parameterized by a prior distribution $p(x_T) = N(x_T; \mathbf{0}, \mathbf{I})$ and a series of learnable transition kernels $p_\theta(x_{t-1}|x_t), t = 1,2,\cdots T$. The prior distribution $p(x_T) = N(x_T; \mathbf{0}, \mathbf{I})$ is chosen because the forward process is constructed such that $q(x_T) \approx N(x_T; \mathbf{0}, \mathbf{I})$. The learnable transition kernels $p_\theta(x_{t-1}|x_t)$ take the form of

$$p_\theta(x_{t-1}|x_t) = N(x_{t-1}; \mu_\theta(x_t, t), \Sigma_\theta(x_t, t)) \quad (3)$$

where the mean $\mu_\theta(x_t, t)$ and variance $\Sigma_\theta(x_t, t)$ are represented by parameterized deep neural networks with $\theta$ denoting the learnable model parameters. The generation of a data sample $x_0$ can be achieved by first sampling a noise vector

$x_T \sim p(x_T) = N(x_T; \mathbf{0}, \mathbf{I})$, and then iteratively applying the learned transition kernel $x_{t-1} \sim p_\theta(x_{t-1}|x_t)$ from $t=T$ to $t=1$ to produce a sequence of samples. This process is repeated until reaching $x_0$.

To ensure the success of this sampling process, it is crucial to train the reverse Markov chain to accurately replicate the time reversal of the forward Markov chain. During the training phase, the parameters $\theta$ should be adjusted to minimize the model's cross-entropy loss, defined as $L_{CE} = -E_{q(x_0)} \log p_\theta(x_0)$.

Through application of Jensen's inequality, it can be deduced that

$$L_{CE} = -E_{q(x_0)} \log p_\theta(x_0) \leq$$
$$E_{q(x_{0:T})} \left[ \log \frac{q(x_{1:T}|x_0)}{p_\theta(x_{0:T})} \right] = -L_{VLB}(x_0) \quad (4)$$

It is observed that the cross-entropy loss is upper-bounded by the negative variational lower bound. Minimizing the cross-entropy loss involves minimizing the negative variational lower bound, which is the objective of training a DDPM. Moreover, the negative variational lower bound can be decomposed into a sum of independent terms [19], allowing for efficient estimation using Monte Carlo sampling.

One simplified objective function to minimize, as given by Ho et al. [21], is

$$L^{simple} = E_{t \sim U[1,T], x_0 \sim q(x_0), \epsilon \sim N(\mathbf{0},\mathbf{I})} [\| \epsilon - \epsilon_\theta(x_t, t) \|^2] =$$
$$E_{t \sim U[1,T], x_0 \sim q(x_0), \epsilon \sim N(\mathbf{0},\mathbf{I})} [\| \epsilon - \epsilon_\theta(\sqrt{\bar{\alpha}_t}x_0 + \sqrt{1-\bar{\alpha}_t}\epsilon, t) \|^2] \quad (5)$$

where $U[1,T]$ is a uniform distribution over the set $\{1,2,...,T\}$. The deep neural network $\epsilon_\theta$ with parameters $\theta$ predicts the noise vector $\epsilon$ given $x_t$ and $t$. The prediction of $\mu_\theta(x_t, t)$ can be achieved using $\epsilon_\theta(x_t, t)$, as shown in the following expression:

$$\mu_\theta(x_t, t) = \frac{1}{\sqrt{\alpha_t}} \left( x_t - \frac{1-\alpha_t}{\sqrt{1-\bar{\alpha}_t}} \epsilon_\theta(x_t, t) \right) \quad (6)$$

## III. METHODOLOGY

### A. Problem Formulation

We consider a population of agents (such as the population of one country), indexed by $n=1, 2, ..., N$, where each agent is characterized by a vector of features, denoted as $x_n$. These features represent the individual characteristics of each agent, such as gender, age, educational background, and so on. Formally, $x$ is a random vector, and each $x_n$ represents a realization of $x$. The population synthesis problem seeks to estimate a joint probability distribution $\hat{p}(x)$ that approximates the true joint probability distribution $p(x)$ of features across a population. Once this distribution is reasonably approximated, a synthetic population of $N$ individuals can be generated through random sampling from $\hat{p}(x)$. Typically, estimation of $\hat{p}(x)$ is performed using a sample of size $M$, which is often two orders of magnitude smaller than $N$. In this study, a customized diffusion model is designed and trained to provide an accurate estimation $\hat{p}(x)$ of the true probability distribution $p(x)$.

As noted in Section I, synthetic population generation models produce both sampling zeros and structural zeros. Fig. 2 presents an Euler diagram illustrating their relationship. The set $X$ encompasses all possible combinations of features, including logically infeasible combinations that do not exist in the real world. The set $X_p$ represents the combinations of features that arise in the real-world population. The set $X_s$ denotes the combinations of features collected through the survey. Finally, the set $X_g$ signifies the generated sample produced by a deep generative model.

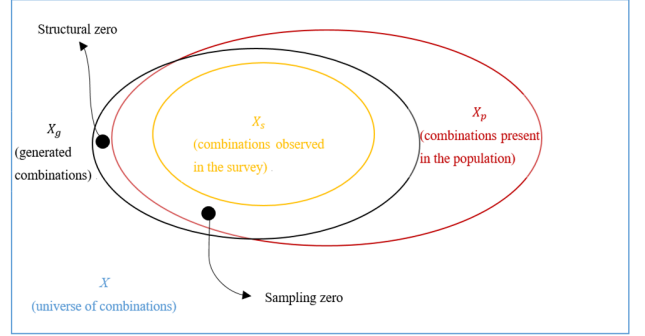

**Fig. 2.** Conceptual diagram of the sampling zeros, the structural zeros and the different sets of combinations that arise in population synthesis.

In this study we present a generative method for population synthesis based on diffusion model, which accurately approximates the true joint probability distribution of features across the population.

### B. Diffusion Model-based Population Synthesis

Most deep generative models employed in population synthesis focus on VAEs and GANs. To our knowledge, only one previous application of the diffusion model to this domain is reported in [33], which involved aligning the individual data with the input format required by the conventional image-generating diffusion model through forcefully reshaping individual sample data into square matrices from vectors. However, this approach did not fully leverage the representation capabilities of the diffusion model for population data. Consequently, we designed and trained a customized diffusion model specifically tailored to approximate the true joint probability distribution of the population, thereby facilitating more effective synthetic pools generation.

The detailed data representation process that converts original discrete attribute combinations into continuous matrices for customized DDPM is as follows. First, the individual sample vector with discrete attribute values is transformed into a matrix by embedding each attribute value as a row vector, analogous to word embeddings in natural language processing. Specifically, a dictionary is constructed by collecting all discrete attribute values from the sample vectors. One-hot encoding is then applied to the original sample vectors based on this dictionary, converting the discrete attribute vectors into continuous matrix inputs for the customized DDPM. To facilitate effective learning and representation, a fully connected layer is introduced as the initial layer of the customized DDPM. The combination of the fully connected learnable embedding layer and the preceding one-hot encoding process enables the effective embedding of the discrete input, resulting in a comprehensive representation that can be further integrated with positional encoding. This is demonstrated in Fig. 3.

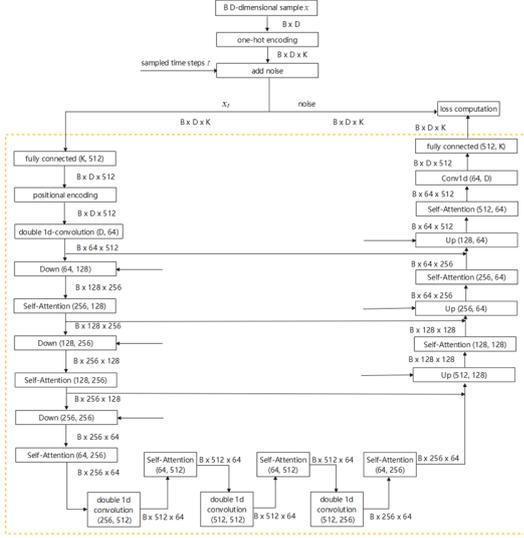

**Fig. 3.** Noise prediction network of the customized DDPM

Fig. 3 illustrates the detailed network architecture used for training the customized DDPM. The core objective of the DDPM training process is to learn a noise prediction network $\epsilon_\theta(x_t, t)$, from the input $x_t$, which predicts the accumulated noise until time step $t$. The area enclosed by the orange dashed box represents the noise prediction network $\epsilon_\theta(x_t, t)$, as previously described in Section II. During training, the true joint probability distribution is indirectly approximated through $\epsilon_\theta(x_t, t)$ by effectively exploring the dependencies among the embeddings, utilizing 1D convolution and multi-head self-attention mechanisms.

Here, we assume that the population data is $D$-dimensional, implying $D$ feature attributes. The dictionary size, denoted as $K$, represents the total number of discrete categorical attribute values across all features. The batch size is denoted as $B$. Additional details on sub-modules of the noise prediction network are provided in Fig. 4.

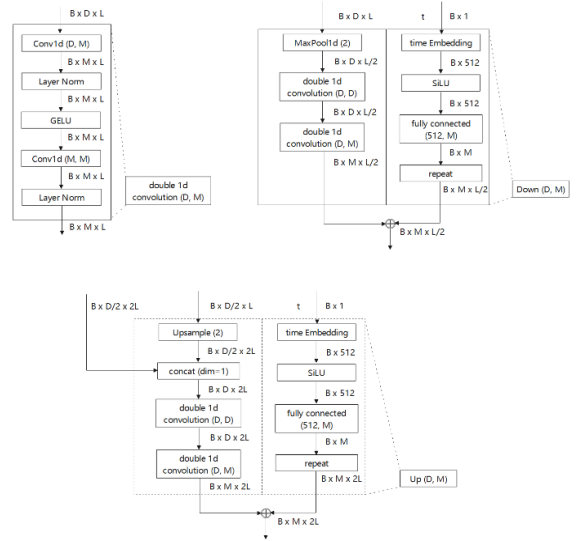

**Fig.4.** The detailed structures of the sub-modules

The training and generation algorithms of the classic DDPM [21] are directly utilized in this study. The primary focus in this study lies in designing a customized diffusion model for generating discrete population data. The training and generation algorithms are depicted in Algorithm 1 and Algorithm 2 below.

**Algorithm 1** Training algorithm
1  **repeat**
2  $x_0 \sim q(x_0)$
3  $t \sim \text{Uniform}(\{1, \cdots, T\})$
4  $\epsilon \sim \mathcal{N}(\mathbf{0}, \mathbf{I})$
5  take gradient descent step on
$$\nabla_\theta \left\| \epsilon - \epsilon_\theta(\sqrt{\bar{\alpha}_t} x_0 + \sqrt{1 - \bar{\alpha}_t} \epsilon, t) \right\|^2$$
6  **until** converged

**Algorithm 2** Generation algorithm
1  $x_T \sim \mathcal{N}(\mathbf{0}, \mathbf{I})$
2  **for** $t = T, \cdots, 1$ **do**
3  $z \sim \mathcal{N}(\mathbf{0}, \mathbf{I})$ if $t > 1$, else $z = \mathbf{0}$
4  $x_{t-1} = \frac{1}{\sqrt{\alpha_t}} \left( x_t - \frac{1 - \alpha_t}{\sqrt{1 - \bar{\alpha}_t}} \epsilon_\theta(x_t, t) \right) + \sigma_t z$
5  **end for**
6  **return** $x_0$

The training algorithm serves as a step-by-step implementation of equation (5). Upon obtaining the noise prediction network $\epsilon_\theta(x_t, t)$ through training on the specified dataset, the generation process can be directly executed by following the steps illustrated in the generation algorithm depicted in Algorithm 2. One important point to note is that the output $x_0$ of a single sample from the generation algorithm is a matrix(tensor) of size $D \times K$. Therefore, it is necessary to perform the $argmax$ operation across dimension 1 and subsequently apply one-hot decoding to obtain the final $D$-dimensional output.

## IV. APPLICATION

In this study, population synthesis experiments are conducted using a large household travel survey (HTS) dataset to demonstrate the superiority of the proposed customized diffusion model in generating synthetic pools. Due to the higher training cost during sampling in the diffusion model compared to other models, optimal hyperparameter tuning was not performed. The default hyperparameters provided in the Keras example for DDPM (https://keras.io/examples/generative/ddpm/) were adopted for training. As mentioned in Section II, the $\beta_t$ parameter, which adds noise to the data during the forward process, was linearly increased from 1e-4 to 0.02 over the total 1000 steps. Model training consists of 700 epochs with an initial learning rate of 3e-4 and utilizes a CosineAnnealingLR scheduler with T_max=700, which decreases the learning rate monotonically to 1e-7 by epoch 700. The AdamW optimization algorithm was used to minimize the loss function. Training was performed on an RTX4080 GPU.

### A. Data Acquisition

#### 1) Data Preparation and Assumptions

The large-scale sample dataset with over 1 million individuals, created by combining household travel survey (HTS) data from South Korea conducted in 2010, 2016 and 2021, is used to address the issue of unavailability of the entire population. We consider the entire dataset as an *h*-population and randomly select only 5% of it as an *h*-sample, comprising approximately 50 thousand individuals. The model is trained using this *h*-sample. The rationale behind this approach is that assuming the *h*-population is representative of the true population, we can expect that the relative diversity of the *h*-sample to the *h*-population mirrors that of the HTS sample (i.e., *h*-population) to the unknown true population, given a similar sampling rate. This expectation allows us to calculate precision and recall metrics [18] (See Section IV for details) to assess feasibility and diversity respectively, which can be applied to the HTS sample of the unknown true population without loss of generality. This scheme is used solely for evaluating the methodology. In the actual generation of the synthetic pools, the entire dataset (the *h*-population) will be used to train the diffusion model.

The HTS sample consists of 1066319 attribute combinations of individuals, which forms the *h*-population. Table I presents descriptive statistics of the *h*-population. It features 13 individual attributes, where all numerical attributes have been discretized into categorical attributes, resulting in a total of 70 categories across these attributes.

To ensure consistency and relevance across the dataset, we adjusted the categories of each attribute to match the data from three different years (2010, 2016, and 2021). The numerical attributes are categorized into distinct intervals. By using only categorical attributes, we can represent the data distribution with a finite number of attribute combinations, which facilitates better evaluation and analysis in subsequent stages.

TABLE I
DESCRIPTIVE STATISTICS OF THE H-POPULATION (N=1066319)

| Attribute | Category | Proportion(%) | Category | Proportion(%) |
|---|---|---|---|---|
| 1.Age(17) | [5,10) | 4.96 | [50,55) | 8.89 |
| | [10,15) | 7.59 | [55,60) | 7.38 |
| | [15,20) | 7.48 | [60,65) | 5.57 |
| | [20,25) | 4.96 | [65,70) | 4.27 |
| | [25,30) | 6.08 | [70,75) | 3.02 |
| | [30,35) | 7.03 | [75,80) | 2.23 |
| | [35,40) | 9.42 | [80,85) | 1.16 |
| | [40,45) | 9.90 | [85,90] | 0.42 |
| | [45,50) | 9.67 | | |
| 2.Sex(2) | Male | 51.23 | Female | 48.77 |
| 3.Household income(5) | <1 million | 8.47 | 5million-10million | 16.09 |
| | 1million-3million | 39.46 | >10 million | 2.19 |
| | 3million-5million | 33.78 | | |
| 4.Hometype(6) | apartment | 55.41 | single house | 21.32 |
| | villa | 12.09 | dual purpose house | 0.82 |
| | multi-family | 9.48 | other | 0.89 |
| 5.Householder car owner(2) | yes | 83.91 | no | 16.09 |
| 6.Driver's license(2) | yes | 60.13 | no | 39.87 |
| 7.Number of working days(4) | 5 days per week | 27.81 | 1-4 days per week | 10.05 |
| | 6 days per week | 17.33 | Inoccupation/non-regular | 44.82 |
| 8.Working types(9) | student | 15.45 | manager/office | 11.54 |
| | inoccupation | 18.40 | agriculture and fisher | 5.68 |
| | experts | 11.07 | simple labor | 12.31 |
| | service | 15.69 | others | 4.43 |
| | sales | 5.44 | | |
| 9.Student(4) | kid | 0.57 | university | 4.72 |
| | Elementary/middle/high | 17.88 | none | 76.83 |
| 10.Number of households(7) | 1 | 7.56 | 5 | 9.67 |
| | 2 | 18.16 | 6 | 1.32 |
| | 3 | 25.27 | 7 | 0.14 |
| | 4 | 37.88 | | |
| 11.Kid in the household(2) | yes | 11.04 | no | 88.96 |
| 12.Commuting mode(6) | car | 25.46 | taxi | 0.31 |
| | bike/bicycle | 2.13 | walking | 21.37 |
| | public transportation | 22.32 | none | 28.40 |
| 13.Departure time of commuting(4) | peak | 56.56 | others | 2.34 |
| | Non-peak | 13.44 | none | 27.66 |

#### 2) Sampling Zeros According to A Sampling Rate

Traditional sampling-based methods fail to generate sampling zeros in population synthesis. To gain insight into this phenomenon, we conducted an evaluation of the degree of sampling zeros based on the sampling rate using the *h*-population and *h*-sample, as shown in Fig. 5.

First, we extracted all unique attribute combinations from the *h*-population and measured the number of unique combinations sampled from the *h*-sample (denoted as 'sampled combinations'). There were only 30820 unique combinations in the *h*-sample, compared to 264005 in the *h*-population, indicating that the sampled unique combinations represented just 11.67% of the total.

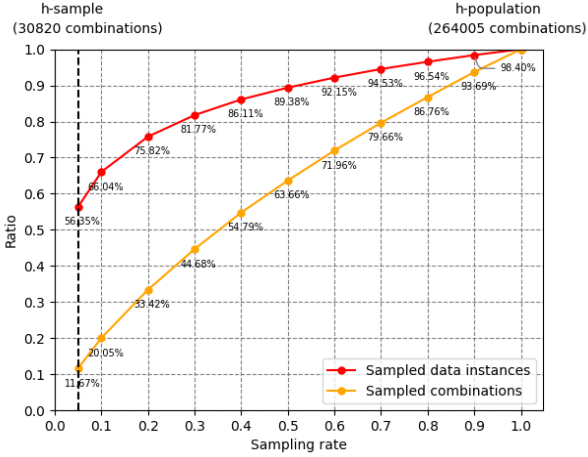

**Fig. 5.** The degree of sampling zeros according to the sampling rate

Second, we examined the ratio of data instances in the population whose attribute combinations were included in the sample at different sampling rates (represented by the red line). This measure, which we define as 'recall' later on, revealed that 56.35% of *h*-population observations or data instances were represented by just 11.67% of sampled combinations.

This result suggests that if population synthesis is conducted using the prototypical agent approach, the diversity of the synthetic population would be limited to the subset of population data instances represented by a mere 11.67% of attribute combinations, which corresponds to 56.4% of the total observations.

*B. Evaluation Metrics*

*1) Distributional Similarity*

Unostentatiously, the diffusion model can be evaluated by comparing the distributional similarity between the actual population distribution and the generated distribution. However, assessing high-dimensional distributions can be computationally expensive and may not always yield consistent results [17]. An alternative approach is to evaluate simplified distributional similarity by examining marginal and bivariate distributions instead [6], [14]. Although this method risks overfitting to marginal and bivariate distributions as noted in [16], it can still serve as a useful indicator of whether the synthetic data generated by the diffusion model is reasonable. We use the standardized root mean square error (SRMSE) to evaluate the distributional similarity of marginal and bivariate distributions [6]. A lower SRMSE indicates a higher level of similarity. The SRMSE for bivariate distributions is specified in Equation (7).

$$\text{SRMSE}(\pi, \hat{\pi}) = \frac{\text{RMSE}(\pi, \hat{\pi})}{\bar{\pi}} = \frac{\sqrt{\sum_{(k,k')}(\pi_{(k,k')} - \hat{\pi}_{(k,k')})^2 / N_b}}{\sum_{(k,k')} \pi_{(k,k')} / N_b} \quad (7)$$

where $\pi$ and $\hat{\pi}$ denote the categorical distributions of the *h*-population and the generated data, respectively. $N_b$ is the total number of category combinations. For the bivariate distribution of categorical variables $k \neq k'$, the vector $\pi = \{\pi(x_1, x_2), \ldots, \pi(x_k, x_{k'})\}$ encompasses all $\binom{K}{2}$ bivariate combinations of the categorical variables. In the case of the marginal distribution, the vector $\pi = \{\pi(x_1), \ldots, \pi(x_K)\}$ represents the marginal distributions for the $K$ categorical variables.

*2) Feasibility and Diversity*

In addition to distributional similarity metrics, we also evaluate the feasibility and diversity of the generated data to provide a comprehensive assessment. This is because sampling zeros and structural zeros are closely related to these aspects. Feasibility measures how closely the generated data resembles the population data. In contrast, Diversity refers to how well the generated data captures the variations present in the population. We evaluate these aspects using precision and recall, respectively.

While diversity assesses the extent to which the generated data overfits the population, the feasibility examines whether the generated data, including new attribute combinations, still accurately reflects the characteristics of the population data. As utilized in [18], the precision and recall for population synthesis, based on the *h*-population ($X$) and the generated data ($\hat{X}$), can be formulated as follows,

$$Precision = \frac{1}{M} \sum_{j=1}^{M} 1_{\hat{X}_j \in X} \quad (8)$$

$$Recall = \frac{1}{N} \sum_{i=1}^{N} 1_{X_i \in \hat{X}} \quad (9)$$

where $M$ is the number of individuals in the generated population and $N$ is the number of individuals in the *h*-population. The function $1_{(\cdot)}$ is an indicator function for counting. To ensure a fair comparison between the generated data and the *h*-population, we assume that $M=N$ in the experiment, i.e. the size of the generated population is equal to the size of the reference *h*-population. Since precision and recall have a trade-off relationship, meaning that optimizing one metric may degrade the other. To address this issue, the F1 score is also calculated to indicate the overall quality of the generated data.

$$\text{F1 score} = \frac{2 \times Precision \times Recall}{Precision + Recall} \quad (10)$$

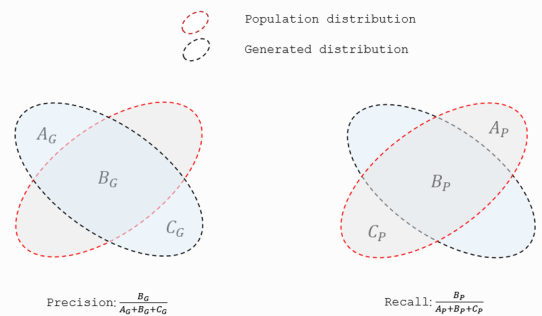

**Fig. 6.** Conceptual diagram of precision and recall for generative model evaluation

Fig.6 illustrates how feasibility and diversity are connected to precision and recall. Precision measures the proportion of the generated data that falls within the *h*-population, which is equivalent to one minus the structural zero rate. Recall measures the proportion of *h*-population combinations that presents in the generated data, reflecting the diversity captured by the generated data.

*C. Results*

Table II summarizes the experimental results of our proposed diffusion model, demonstrating its superior performance compared to established deep generative models (WGAN and VAE) as reported in [18]. This study [18] had previously achieved strong results on the HTS dataset. Due to the challenges associated with replicating the original author's methodology, we have opted to directly cite the experimental results from the original paper and compare them to those obtained using our proposed approach.

TABLE II
EXPERIMENTAL RESULTS OF GENERATED DATA WITH SIZE OF H-POPULATION

| Method | | | Distributional Similarity | | Diversity | | Feasibility | Overall Quality |
|---|---|---|---|---|---|---|---|---|
| Model | Space | Loss functions | Marg. SRMSE | Bivar. SRMSE | # of comb* | Recall | Precision | F1 score |
| VAE | Discrete | $R_{BD}$ | 0.095 | 0.218 | 265875 | 79.9% | 79.2% | 79.5% |
| | | $R_{AD}$ | 0.057 | 0.132 | 312506 | 82.1% | 76.1% | 79.0% |
| | | $R_{BD}\&R_{AD}$ | 0.079 | 0.173 | 329097 | 82.0% | 73.63% | 77.6% |
| | Embedded | $R_{BD}$ | 0.088 | 0.208 | 289377 | 80.2% | 76.6% | 78.4% |
| | | $R_{AD}$ | 0.060 | 0.140 | 334569 | 82.3% | 74.1% | 78.0% |
| | | $R_{BD}\&R_{AE}$ | 0.050 | 0.116 | 318731 | 82.0% | 75.4% | 78.6% |
| WGAN | Discrete | $R_{BD}$ | 0.036 | 0.094 | 155586 | 74.7% | 89.0% | 81.2% |
| | | $R_{AD}$ | 0.016 | 0.048 | 273622 | 81.2% | 80.4% | 80.8% |
| | | $R_{BD}\&R_{AD}$ | 0.043 | 0.106 | 152031 | 74.1% | **89.2%** | 81.0% |
| | Embedded | $R_{BD}$ | 0.023 | 0.076 | 225408 | 77.7% | 84.6% | 81.0% |
| | | $R_{AD}$ | 0.020 | 0.059 | 276012 | 81.3% | 80.3% | 80.8% |
| | | $R_{BD}\&R_{AE}$ | 0.024 | 0.072 | 236238 | 78.1% | 83.0% | 80.5% |
| Diffusion Model | Embedded | MSE | **0.014** | **0.034** | 288862 | **83.8%** | 81.1% | **82.4%** |

*Note*: **Bold** font indicates the best model for each metric. # of comb indicates the number of unique combinations of the generated data. The subset of the population used for training our proposed method comprises 53506 samples, featuring 30820 unique combinations.
*: The number of combinations of population data is 264005

*1) Distributional Similarity*

The distributional similarity between the population data and generated data is quantified using the SRMSE metric, which assesses both marginal and bivariate distributions. Although this approach can be prone to overfitting the marginal and bivariate distributions, as noted in [16], it remains a useful indicator for verifying whether deep generative models have been properly trained.

The results in Table II demonstrate that our proposed diffusion model outperforms both VAE and WGAN models in terms of distributional similarity, achieving the lowest SRMSE values for both marginal and bivariate distributions. Notably, this performance is achieved without customizing loss functions or employing complex preprocessing techniques; instead, the common MSE loss function is used to train the noise prediction network, while simple one-hot embedding coding is applied to discrete inputs. This promising outcome underscores the robustness and effectiveness of our proposed diffusion model.

Fig. 7 visualizes the marginal distributional similarity between the generated samples from our proposed diffusion model and the *h*-population, confirming from this perspective that the model has been successfully trained.

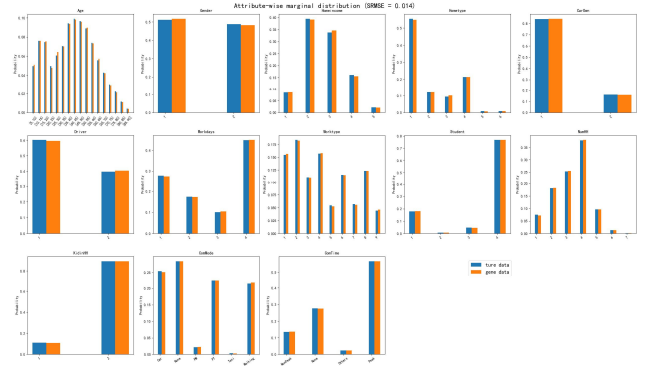

**Fig. 7.** Distributional similarity between the *h*-population and diffusion model generated samples

There are 78 bivariate distributions for all possible attribute pair combinations. As illustrations, here we only display one set of joint distributions. Fig. 8 shows the joint distribution of age and home income in the diffusion model. The left panel and the middle panel show the joint distributions in the *h*-population and the synthesized data, respectively. The right panel displays the fit between the two sets of data.

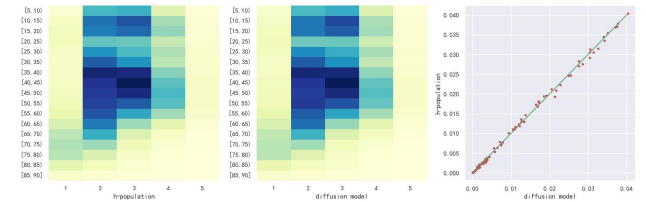

**Fig. 8.** Joint distribution of age and home income

Table II reveals that all variants of WGAN outperform their VAE counterparts in terms of distributional similarity. A possible explanation for this disparity is that WGAN generates fewer sample combinations than VAE, resulting in lower SRMSE values due to reduced structural zeros. This finding highlights the limitations of relying solely on distributional similarity metrics to evaluate the effectiveness of population synthesis methods. To gain a more comprehensive understanding, we additionally investigate the diversity and feasibility of our proposed diffusion model in the following subsections.

*2) Diversity Measured by Recall*

The diversity metric, measured by recall, indicates the proportion of population data whose attribute combinations are captured by the generated data. This metric reflects the ability of the model to generate novel samples that deviate from the training data, closely tied to the concept of sampling zero. In our proposed diffusion model, we obtained 288862 unique attribute combinations in generating over one million (1066319) sample data using just 53506 randomly selected training samples from the $h$-population dataset. As shown in Table II, our model outperforms both WGAN and VAE variants in terms of diversity.

When all three types of deep generative models (VAE, WGAN and diffusion model) produce an equal number of data samples, matching the size of the $h$-population, VAE variants typically excel their WGAN counterparts and diffusion model in terms of generating unique attribute combinations on average. The number of generated unique attribute combinations is related to recall, but not directly proportional. Across all variants, VAE models consistently surpass corresponding WGAN models in terms of diversity. While the proposed diffusion model generates fewer unique attribute combinations compared to VAE variants, it achieves a superior diversity metric, indicating that it excels at producing sampling zeros. In contrast, VAE variants tend to generate more structural zeros.

*3) Feasibility Measured by Precision*

The feasibility of a synthetic population, measured by precision, reflects the proportion of generated data whose attributes exist in the population data. Minimizing infeasible synthetic population is crucial for ensuring the reliability of population synthesis, but this goal often comes at the expense of sacrificing generation diversity. According to Table II, WGAN variants excel at producing realistic data, achieving higher precision than the other two models. However, this strength comes at a cost: they generate significantly fewer unique attribute combinations, resulting in a relatively low diversity metric. In contrast, our proposed diffusion model achieves a better balance between feasibility and diversity.

Our model's feasibility metric is lower than WGAN's but higher than VAE's. Moreover, its superior diversity metric compared to WGAN and VAE suggests that it has achieved a desirable tradeoff between diversity and feasibility.

*4) Overall Quality*

To reconcile the trade-off between feasibility and diversity, we use the overall quality (F1 score) as the primary metric for evaluating our model. As shown in Table II, our proposed diffusion model excels as the top-performing model, achieving an F1 score of 82.4% with recall and precision rates of 83.8% and 81.1% respectively. This impressive performance highlights the proposed diffusion model's ability to strike a better balance between feasibility and diversity.

V. CONCLUSION

This study aims to enhance the initial stage of population synthesis, a crucial step in generating input data for agent-based transportation simulation. The objective is to create synthetic pools of individuals that reflect the diverse attribute combinations present in real-world population. To achieve this goal, a novel customized diffusion model is proposed, with a specific focus on household travel data generation. This approach leverages the strengths of diffusion models to produce high-quality, realistic population data that can be effectively utilized in various applications.

The proposed customized diffusion model efficiently approximates the joint probability density function of attributes by embedding discrete attribute combinations as continuous vectors and leveraging self-attention modules. This enables the generation of high-quality samples that capture the complexities of real-world population.

When evaluating the proposed diffusion model, we go beyond traditional marginal distribution similarity metrics by computing additional key performance indicators: diversity and feasibility. Diversity is quantified as the proportion of the population whose attributes are covered by the generated data, ensuring that the synthetic population accurately represents the real-world demographic landscape. Feasibility measures the proportion of the generated data whose attributes exist in the population data, guaranteeing that the synthetic populations align with empirical reality. Therefore, a comprehensive assessment to the proposed method can be achieved by taking all these metrics into consideration.

In summary, our research makes two significant contributions to the field of population synthesis. Firstly, we have successfully demonstrated the applicability and effectiveness of diffusion models in generating high-quality population data. While deep learning methods such as VAEs and GANs have been extensively explored for synthetic pools generation, our work is the first to thoroughly investigate the use of diffusion models for this purpose. Secondly, by designing a customized diffusion model specifically tailored for handling discrete attribute combinations, we achieve better results on the HTS dataset compared to VAEs and WGANs. Our proposed diffusion model not only improves marginal distribution similarity measures but also enhances diversity metric while maintaining a better balance between diversity and feasibility. These improvements unequivocally demonstrate the effectiveness of our approach in generating accurate population data.

Throughout this research, several new and intriguing research directions have emerged that merit further investigation.

To begin with, the generative speed of our proposed diffusion model can be accelerated to boost computational efficiency. The classic DDPM training and generation algorithms were employed in our method for simulating both the forward and inverse diffusion processes. Traditional DDPM approach requires an excessively large number of steps (>=1,000) to convert noise samples into actual data points, resulting in computationally intensive generation processes. To address these performance concerns, alternative diffusion models such as analytic-DPM [36] and DDIM [37] can be explored as potential substitutes. These alternative approaches may offer reduced generation times, making them particularly suitable for applications where real-time or near-real-time population synthesis is critical.

Another area worth exploring is conditional generation of synthetic populations via diffusion models. In most practical applications of diffusion models, conditioning on specific constraints is inherent. For example, generating images based on descriptive texts, where the texts serve as the conditioning information. However, the ability to generate synthetic populations that conform to specific marginal distributions using diffusion models has not been explored, despite its importance in real-world applications.